\documentclass[conference]{IEEEtran}
\usepackage{times}

\usepackage[numbers,sort&compress,square]{natbib}
\usepackage{multicol}
\usepackage[bookmarks=true]{hyperref}
\usepackage{pdfpages}
\usepackage{float}
\usepackage{balance}

\IEEEoverridecommandlockouts
\begin{document}
\raggedbottom

\title{Socially intelligent task and motion planning for human-robot interaction}



\author{\parbox{18cm}{\centering
     Andrea Frank  and Laurel D. Riek\thanks{Some research reported in this article is based upon work supported by the National Science Foundation under Grant IIS-1527759.}\\
     Computer Science and Engineering, UC San Diego
     \thanks{Presented at RSS 2019 Workshop on Robust Task and Motion Planning.}
     \\ 
    }
     }

\maketitle

\IEEEpeerreviewmaketitle

\section{Introduction}



Robots are entering human social environments (HSEs) such as hospitals, homes, and factories \cite{riek2017healthcare}.
As social beings,
much human behavior is predicated on \textit{social context} - the ambient social state that describes cultural norms, social signals, and individual preferences \cite{riek2014social,nigam2015social,o2015detecting}.
Cues such as facial expressions, body posture, and group behavior encode social affordances that are critical for understanding and predicting human behavior \cite{hall1968proxemics,hogan2003can,moosaei2017using,taylor2019coordinating}.
Furthermore, humans consistently view, communicate with, and react to robots as social actors \cite{reeves1996media,breazeal2004social,sardar2012don,baxter2018robots,cunningham2019mpdm}.
Many tasks can only be completed by understanding, respecting, and interacting with social context.
For example, even in the simple task of buying an ice cream, one must understand why people line up to order, wait in line without cutting, and execute an interactive dialogue before receiving the treat.
Thus, in order to act both appropriately and effectively in HSEs, 
it is crucial for robots to understand social context and how to leverage social interactions to complete tasks.


Endowing robots with social agency---the understanding of oneself as a social actor---is an open challenge that must be addressed before robots can plan safe and reliable behavior in HSEs \cite{riek2014social,nigam2015social,o2015detecting,charalampous2017recent,riek2017healthcare}.
Socially-aware planning seeks to model social context in order to generate appropriate policies for HSEs \cite{khambhaita2017assessing, charalampous2017recent, rizwan2018human,cunningham2019mpdm}.
Many existing socially-aware planners use proxemics, the social affordance of space \cite{hall1968proxemics}, to represent social context as an objective function over the environment \cite{mumm2011human,rios2015proxemics,okal2016formalizing}.
These planners consider humans as \textit{social obstacles}, i.e. obstacles with an additional social occupancy that encodes proxemic affordances such as personal space \cite{bera2017sociosense} and visibility preferences \cite{kirby2009companion}.
Social obstacle planners then generate human-aware paths on these costmaps using lightweight graph-search \cite{okal2016formalizing}, potential field \cite{huang2010human}, or sampling-based \cite{svenstrup2010trajectory} planners.
However, these path planners do not consider high-level decisions about subtasks beyond motion and obstacle avoidance, and are therefore insufficient for planning the complex, interactive policies necessary in HSEs.

Joint task and motion planning (TMP) is a central area of research in robotics and artificial intelligence that seeks to address this issue by sharing information between task and motion planners.
Many TMP algorithms interleave task and motion planning, replanning at the task level (i.e. reordering or choosing new subtasks) when the high-level symbolic plan violates low-level geometric constraints (e.g. the robot is unable to reach the desired object without collision) \cite{kaelbling2013integrated,dantam2016incremental,garrett2018ffrob,srivastava2014combined}.
The majority of this work focuses on  manipulation tasks, where the interactions between the robot and the world are simple and well-defined.
Few studies have addressed joint TMP in HSEs \cite{pellegrinelli2016probabilistic,pellegrinelli2017motion}, and these approaches still largely incorporate human models for the purpose of avoiding collisions during task execution.
While these methods are a great start to a difficult problem, one key gap is the consideration of social context.
Without this, robots are at risk of both distressing those around it by violating social norms and of failing to find the optimal task plan by ignoring possible social interactions.

It is nontrivial to integrate social awareness with modern TMP.
Social constraints can be considered relaxed constraints, i.e. constraints that can be only partially satisfied when needed to meet a stronger constraint such as collision avoidance.
While some researchers incorporated relaxed constraints into robot planners \cite{kunz2012manipulation,stilman2010global,huh2018constrained}, the majority only do so in the context of replanning low-level obstacle avoidance policies to satisfy a static task.
The few works that explore relaxed constraints in dynamic TMP either only replan subtask order \cite{lahijanian2016iterative,guo2018probabilistic} or consider only simple, well-defined subtasks (e.g. recording information, picking up a cup) \cite{lahijanian2016iterative,ceriani2015reactive,guo2015multi}.

Furthermore, most existing work does not weight the degree of strictness of different constraints.
While \cite{ceriani2015reactive} and \cite{guo2015multi} propose ways to weight constraints, these methods rely on a human expert to designate their importance.
In HSEs, the relative importance of constraints varies with social context, and since social context is also influenced by each robot action, it is inadmissible to assume static, externally-supplied weights.
To our knowledge, there exists no joint task and motion planner that is able to reason about complex high-level task replanning with dynamically-varying, relaxed constraints.

\section{Proposed Algorithm}

In this work, we propose a socially-aware TMP algorithm that leverages an understanding of social context to generate appropriate and effective policy in HSEs.
The key strength of our algorithm is that it explicitly models how potential actions affect not only objective cost (e.g. path length), but transform social context (e.g. interrupting conversations via an alarm).
We also inform the planner of the relative importance or urgency of its current task goal, which it uses along with its own calculation of social constraints to determine when it is and is not appropriate to violate social expectations to optimize the objective function \cite{nigam2015social}.
This social awareness allows a robot to understand a fundamental aspect of society: just because something makes your job easier does not make it the right thing to do.

We make several assumptions in this work. We assume knowledge of the location of the robot, static obstacles, and people. We further assume that there is a finite, known set of available interactions the robot can initiate with a person or group of people. We do not assume perfect knowledge of social context, and instead estimate its state from observing human activity and social cues. From our previous work in real-world hospitals and factories, we found it is feasible to employ wearable sensors to accurately track people and equipment  and recognize important activities \cite{kubota2019activity,frank2019wearable}.

In this work, we adopt a classic TMP architecture with a high-level task planner and a low-level motion planner.
We separate our robot's social interactions into two domains: active (e.g. asking for help) and passive (e.g. accommodating personal space).
We incorporate passive social interactions into our low-lever planner in a lightweight social obstacle planner in order to leverage the benefits of existing work.
Our task planner models active social interactions based on a partially-observable Markov Decision Process (POMDP).
Here, social context consists of hidden states (e.g. each person's current activity) that are approximated from features such as proxemics \cite{rios2015proxemics} and group behavior \cite{taylor2016robot}.

To choose relevant features for a specific application, we plan to consult experts in the area, extract features via learning algorithms like neural networks, and perform extensive cross-validation of our selection.
We will collect data on social interactions in real-world manufacturing and emergency department settings to accurately model how human-human and human-robot interactions transform social context in these HSEs.
We will then develop interaction models for a set of possible human-robot interactions, which our task planner will use to plan optimal high-level policy.

The high complexity of HSEs poses concerns for POMDPs. 
In order to remain tractable, our algorithm will employ a policy-switching architecture that balances the benefits of sophisticated social reasoning with the speed of a socially-naive motion planner.
In this model, an executor manages the cooperation between the complex task-level POMDP and the lightweight social obstacle motion planner.
The executor uses a lazy approach that activates the high-level planner only where there is significant potential for active social interaction.
Assuming that interaction is only possible in a small subset of states (often those proximal to a human), our algorithm avoids high-level replanning for the majority of states.
Furthermore, this constraint limits the effective state space of the POMDP to this subset.
We anticipate these strategies will enable our algorithm to remain tractable in real-world HSEs.

\section{Evaluation}

To validate our algorithm, we will simulate a crowded emergency department in which a robot must deliver medicine to a patient's room.
We will investigate three scenarios:
1) social interaction is helpful, but unnecessary to complete the task;
2) social interaction is necessary; and
3) social interaction is helpful, but socially unacceptable (e.g. if it would interrupt a clinician treating a patient).

We will perform our study at a medical simulation center at our institution.
We will recruit clinicians and clinical learners to realistically populate the simulation, who will be assigned tasks to simulate various social states.
For each trial, we will record the executed plan's path length, duration, and ``social burden" as estimated by a post-trial questionnaire.
We will compare our algorithm against two competing state-of-the-art methods: a task-naive social motion planner, such as a pure social obstacle model, and a socially-naive TMP, such as \textit{Ffrob} \cite{garrett2018ffrob}.
We anticipate that our algorithm will generate lower cost plans than the task-naive social obstacle planner by leveraging social interaction.
In addition, we expect that the task-naive planner will fail when social interaction is necessary, as in (2) and (3), while our algorithm will find a viable plan.
We also expect that our algorithm will guarantee socially acceptable policy, where the socially-naive TMP may fail in (3).

\section{Discussion}

We anticipate this work will offer the following contributions: 
First, we introduce a socially intelligent joint task and motion planner that generates appropriate and effective policies for robots in HSEs.
Second, we present a framework for dividing social interactions into active and passive domains to limit computational complexity and leverage existing work in social obstacle planning.
Third, we present a socially-aware, lazy policy-switching architecture that selectively activates high-level replanners only when necessary in order to remain effective and tractable in complex HSEs.

To our knowledge, the proposed work is the first TMP algorithm that enables a robot to act as an intelligent social agent in HSEs.
By explicitly modeling social context while remaining tractable in complex environments, we anticipate our approach will be applicable in a variety of real-world applications.
Through this work, robots will be able to understand, respect, and leverage social context to produce acceptable and functional policy in HSEs.

\balance
\bibliographystyle{condensedIEEEtran}
\bibliography{references}

\end{document}